\documentclass[preprint,journal]{vgtc}       % preprint (journal style)
\ifpdf%                                % if we use pdflatex
  \pdfoutput=1\relax                   % create PDFs from pdfLaTeX
  \usepackage{graphicx}                % allow us to embed graphics files
  \DeclareGraphicsExtensions{.pdf,.png,.jpg,.jpeg} % for pdflatex we expect .pdf, .png, or .jpg files
\else%                                 % else we use pure latex
  \usepackage{graphicx}                % allow us to embed graphics files
  \DeclareGraphicsExtensions{.eps}     % for pure latex we expect eps files
\fi%

\graphicspath{{figures/}{pictures/}{images/}{./}{figs/}} % where to search for the images

\usepackage{microtype}                 % use micro-typography (slightly more compact, better to read)
\PassOptionsToPackage{warn}{textcomp}  % to address font issues with \textrightarrow
\usepackage{textcomp}                  % use better special symbols
\usepackage{mathptmx}                  % use matching math font
\usepackage{times}                     % we use Times as the main font
\usepackage{cite}                      % needed to automatically sort the references
\usepackage{tabu}                      % only used for the table example
\usepackage{booktabs}                  % only used for the table example
\usepackage{float}
\usepackage{xcolor}
\usepackage{subfigure}
\usepackage{nicefrac,xfrac}
\usepackage{algorithm}
\usepackage{amssymb}
\usepackage{amsmath}
\usepackage{soul}
\usepackage{censor}
\usepackage{hyperref}
\usepackage[hang,flushmargin]{footmisc} 
\newcommand{\squishlist}{
	\begin{list}{$\bullet$}
		{ \setlength{\itemsep}{0pt}      \setlength{\parsep}{3pt}
			\setlength{\topsep}{3pt}       \setlength{\partopsep}{0pt}
			\setlength{\leftmargin}{1.5em} \setlength{\labelwidth}{1em}
			\setlength{\labelsep}{0.5em} } }
	
	\newcommand{\squishlisttwo}{
		\begin{list}{$\bullet$}
			{ \setlength{\itemsep}{0pt}    \setlength{\parsep}{0pt}
				\setlength{\topsep}{0pt}     \setlength{\partopsep}{0pt}
				\setlength{\leftmargin}{2em} \setlength{\labelwidth}{1.5em}
				\setlength{\labelsep}{0.5em} } }
		
		\newcommand{\squishend}{
		\end{list}  }

\onlineid{1289}

\vgtccategory{Research}
\vgtcpapertype{algorithm/technique}

\title{Explainable Matrix -- Visualization for Global and Local Interpretability of Random Forest Classification Ensembles}

\author{M\'ario Popolin Neto and Fernando V. Paulovich, \textit{Member, IEEE}}
\authorfooter{
\item
  M. Popolin Neto is with Federal Institute of S\~ao Paulo (IFSP) and University of S\~ao Paulo (USP), Brazil. E-mail: mariopopolin@ifsp.edu.br
\item
  F. V. Paulovich is with Dalhousie University, Canada, and University of S\~ao Paulo (USP), Brazil. E-mail: paulovich@dal.ca}

\shortauthortitle{Popolin{ }Neto, M. and Paulovich, F. V.: Explainable Matrix}
\abstract{Over the past decades, classification models have proven to be essential machine learning tools given their potential and applicability in various domains. In these years, the north of the majority of the researchers had been to improve quantitative metrics, notwithstanding the lack of information about models' decisions such metrics convey. This paradigm has recently shifted, and strategies beyond tables and numbers to assist in interpreting models' decisions are increasing in importance. Part of this trend, visualization techniques have been extensively used to support classification models' interpretability, with a significant focus on rule-based models.  Despite the advances, the existing approaches present limitations in terms of visual scalability, and the visualization of large and complex models, such as the ones produced by the Random Forest (RF) technique, remains a challenge. In this paper, we propose \textit{Explainable Matrix (ExMatrix)}, a novel visualization method for RF interpretability that can handle models with massive quantities of rules. It employs a simple yet powerful matrix-like visual metaphor, where rows are rules, columns are features, and cells are rules predicates, enabling the analysis of entire models and auditing classification results. ExMatrix applicability is confirmed via different examples, showing how it can be used in practice to promote RF models interpretability.   
} % end of abstract

\keywords{Random forest visualization, logic rules visualization, classification model interpretability, explainable artificial intelligence}

\vgtcinsertpkg

\begin{document}

%% The ``\maketitle'' command must be the first command after the
%% ``\begin{document}'' command. It prepares and prints the title block.

%% the only exception to this rule is the \firstsection command
\firstsection{Introduction}

\maketitle

% %% \section{Introduction} %for journal use above \firstsection{..} instead

Imagine a machine learning classification model for cancer prediction with $99\%$ accuracy, prognosticating positive breast cancer for a specific patient. Even though we are far from reaching such level of precision, we (researchers, companies, among others) have been trying to convince the general public to trust classification models, using the premise that machines are more precise than humans~\cite{doi:10.1177/117693510600200030}. However, in most cases, yes or no are not satisfactory answers. A doctor or patient inevitably may want to know why positive? What are the determinants of the outcome? What are the changes in patient records that may lead to a different prediction? Although standard instruments for building classification models, quantitative metrics such as accuracy and error cannot tell much about the model prediction, failing to provide detailed information to support understanding~\cite{Liu:2018:Visual}. 

We are not advocating against machine learning classification models, since there is no questioning about their potential and applicability in various domains~\cite{Endert:2017:TheState, Butler:2018:Machine}. The point is the acute need to go beyond tables and numbers to understand models' decisions, increasing trust in the produced results. Typically, this is called model interpretability and has become the concern of many researchers in recent years~\cite{Yang:2019:Evaluating, Carvalho:2019:Machine}. Model interpretability is an open challenge and opportunity for researchers~\cite{Endert:2017:TheState} and also a government concern, as the \textit{European General Data Protection Regulation} requires explanations about automated decisions regarding individuals~\cite{Liu:2018:Interpretable, Carvalho:2019:Machine, Guidotti:2018:Survey}.

Model interpretability strategies are typically classified as global or local approaches. Global techniques aim at explaining entire models, while the local ones give support for understanding the reasons for the classification of a single instance~\cite{Du:2018:Techniques, Carvalho:2019:Machine}. In both cases, interpretability can be attained using inherent interpretable models such as Decision Trees, Rules Sets, and Decision Tables~\cite{decisiontable}, or through surrogates, where black-box models, like Artificial Neural Networks or Support Vector Machines, are approximated by rule-based interpretable models~\cite{Guidotti:2018:Survey, Carvalho:2019:Machine}. The key idea is to transform models into logic rules, using them as a mechanism to enable the interpretation of a model and its decisions~\cite{Lei:2018:Understanding, Guidotti:2018:Local, Castro:2019:Surrogate, Ming:2019:RuleMatrix, Ribeiro:2018:Anchors}.

Recently, visualization techniques have been used to empower the process of interpreting rule-based classification models, particularly Decision Tree models~\cite{Castro:2019:Surrogate, Zhao:2019:iForest, Elzen:2011:BaobabView, Schulz:2011:Treevis}. In this case, given the inherent nature of these models, the usual adopted visual metaphors focus on revealing tree structures, such as the node-link diagrams~\cite{Graham:2009:Asurvey, Zhao:2019:iForest, Ming:2019:RuleMatrix}. However, node-link structures are limited when representing logic rules~\cite{Freitas:2014:Comprehensible, Huysmans:2011:Anempirical, Lima:2009:Domain}, and present scalability issues, supporting only small models with few rules~\cite{Graham:2009:Asurvey, Schulz:2011:Thedesign, Zhao:2019:iForest}. Matrix-like visual metaphors have been used~\cite{Ming:2019:RuleMatrix, Castro:2019:Surrogate} as an alternative, but visual scalability limitations still exist, and large and complex models cannot be adequately visualized, such as the Random Forests~\cite{Breiman:2001:RandomForest, Biau:2016:Arandom}. Among rule-based models, Random Forests is one of the most popular techniques given their simplicity of use and competitive results~\cite{Biau:2016:Arandom}. However, they are very complex entities for visualization since multiple Decision Trees compose a model, and, although attempts have been made to overcome such a hurdle~\cite{Zhao:2019:iForest}, the visualization of entire models is still an open challenge.

In this paper, we propose \textit{Explainable Matrix (ExMatrix)}, a novel method for Random Forest (RF) interpretability based on the visual representations of logic rules. ExMatrix supports global and local explanations of RF models enabling tasks that involve the overview of models and the auditing of classification processes.  The key idea is to explore logic rules by demand using matrix visualizations, where rows are rules, columns are features, and cells are rules predicates. ExMatrix allows reasoning on a considerable number of rules at once, helping users to build insights by employing different orderings of rules/rows and features/columns, not only supporting the analysis of subsets of rules used on a particular prediction but also the minimum changes at instance level that may change a prediction. Visual scalability is addressed in our solution using a simple yet powerful compact representation that allows for overviewing entire RF models while also enables focusing on specific parts for details on-demand. In summary, the main contributions of this paper are:

\squishlist %\begin{itemize}
    \item A new matrix-like visual metaphor that supports the visualization of RF models;
    \item A strategy for Global interpretation of large and complex RF models supporting model overview and details on-demand; and
    \item A strategy to promote Local interpretation of RF models, supporting auditing models' decisions. 
\squishend %\end{itemize}

\section{Related Work}

Typically, visualization techniques aid in classification tasks in two different ways. One is on supporting parametrization and labeling processes aiming to improve model performance~\cite{Ankerst:1999:Visual, Teoh:2003:PaintingClass, Do:2007:Towards, Elzen:2011:BaobabView, Talbot:2009:EnsembleMatrix, Hoferlin:2012:Interactive, Lee:2016:AnInteractive, Liu:2018:Visual}. The other is on understanding the model as a whole or the reasons for a particular prediction. In this paper, our focus is on the latter group, usually named model interpretability. 

Interpretability techniques can be divided into pre-model, in-model, or post-model strategies, regarding support to understand classification results before, during, or after the model construction~\cite{Carvalho:2019:Machine}. Pre-model strategies usually give support to data exploration and understanding before model creation~\cite{Paiva:2015:AnApproach,Choo:2010:iVisClassifier,Migut:2010:Visual,Carvalho:2019:Machine}. In-model strategies involve the interpretation of intrinsically interpretable models, such as Decision Trees, and post-model strategies concerns interpretability of complete built models, and they can be model-specific~\cite{Rauber:2017:Visualizing, Wu:2018:Beyond} or model-agnostic~\cite{Castro:2019:Surrogate, Ming:2019:RuleMatrix, Ribeiro:2018:Anchors, Guidotti:2018:Local}. Both in-model and post-model approaches aim to provide interpretability by producing global and/or local explanations~\cite{Du:2018:Techniques}.  

\subsection{Global Explanation}

Global explanation techniques produce overviews of classification models aiming at improving users' trust in the model~\cite{Ribeiro:2016:Why}. For inherently interpretable models, the global explanation is attained through visual representations of the entire model. For more complex non-interpretable black-box models, such as Artificial Neural Networks or Support Vector Machines, interpretability can be attained through a surrogate process where such models are approximated by interpretable ones~\cite{Ming:2019:RuleMatrix, Castro:2019:Surrogate, Hall:2018:Art}. Decision Trees~\cite{Breiman:1984:Classification, Tan:2005:Introduction, Yin:2014:Fifty} are commonly used as surrogate models~\cite{Castro:2019:Surrogate, Hall:2018:Art}, and whether a surrogate or a classification model per se, the most common visual metaphor for global explanation is the node-link~\cite{Ming:2019:RuleMatrix, Zhao:2019:iForest}, such as the BaobaView technique~\cite{Elzen:2011:BaobabView}. The node-link metaphor's problem is scalability~\cite{Graham:2009:Asurvey, Schulz:2011:Thedesign, Zhao:2019:iForest}, mainly when it is used to create visual representations for Random Forests, limiting the model to be small in number of trees~\cite{Stiglic:2006:Using}. Creating a scalable visual representation for an entire Random Forest model, presenting all decision paths (root node to leaf node paths), remains a challenge even with a considerably small number of trees~\cite{Liu:2018:Visual}.

Although the node-link metaphor is the straightforward representation for Decision Trees, logic rules extracted from decision paths have also been used to help on interpretation~\cite{Lima:2009:Domain}. Indeed, disjoint rules have shown to be more suitable for user interpretation than hierarchical representations~\cite{Lakkaraju:2016:Interpretable}, and a user test comparing the node-link metaphor with different logic rule representations, showed that Decision Tables~\cite{decisiontable}  (rules organized into tables) offers better comprehensibility properties~\cite{Freitas:2014:Comprehensible, Huysmans:2011:Anempirical}. Nonetheless, this strategy uses text for representing rules having as drawback model size~\cite{Freitas:2014:Comprehensible}. Similarly to Decision Tables, our method does not lean on the hierarchical property of Decision Trees. However, instead of using text to represent logic rules, we used a matrix-like visual metaphor, where rows are rules, columns are features, and cells are rules predicates, capable of displaying a much larger number of rules than the textual representations.

The idea of using a matrix metaphor to present rules is not new~\cite{Castro:2019:Surrogate, Ming:2019:RuleMatrix}, and it has been used before by the RuleMatrix technique~\cite{Ming:2019:RuleMatrix}. RuleMatrix is a model-agnostic approach to induce logic rules from black-box models, presenting rules in rows, features in columns, and predicates in cells using histograms. As data histograms require a certain display space to support human cognition, the number of rules displayed at once is reduced. Therefore, not being able to present entire or even parts of Random Forest models (notice that their focus is the visualization of surrogate rules,  not models). Our approach also uses a matrix metaphor; however, we employ a simpler icon (colored rectangular shape) for the matrix cells,  mapping different rule properties (e.g., predicates, class, and others), considerably improving the scalability of the visual representation. Besides the recognized scalability of matrix visualization and custom cells~\cite{Alsallakh:2014:Visualizing, Behrisch:2016:Matrix, Alper:2013:Weighted }, rows and columns order is an important principle~\cite{Wu:2008:MatrixVisualization, Chen:2002:GAP, Chen:2004:MatrixVisualization, Behrisch:2016:Matrix}, and in our approach rules and features can be organized using different criteria, promoting analytical tasks not supported by the RuleMatrix, such as the holistic analysis of Random Forest models through complete overviews. Worthy mentioning that different from usual matrix visual metaphors for trees and graphs that focus on nodes~\cite{Behrisch:2016:Matrix, Graham:2009:Asurvey}, our approach focus on decision paths, which is the object of analysis on Decision Trees~\cite{Lima:2009:Domain, Freitas:2014:Comprehensible, Huysmans:2011:Anempirical}, so representing a different concept.

\subsection{Local Explanation}

Unlike the model overview of global explanations, local explanation techniques focus on a particular instance classification result~\cite{Ribeiro:2018:Anchors, Zhao:2019:iForest}, aiming to improve users' trust in the prediction~\cite{Ribeiro:2016:Why}. As in global strategies, local explanations can be provided using inherently interpretable models or using surrogates of black-boxes~\cite{Ribeiro:2018:Anchors, Guidotti:2018:Local, Strumbelj:2010:AnEfficient}. In general, local explanations are constructed using the logic rule applied to classify the instance along with its properties (e.g., coverage, certainty, and fidelity), providing additional information for prediction reasoning~\cite{Ming:2019:RuleMatrix, Lakkaraju:2016:Interpretable}. 

One example of a visualization technique that supports local explanation is the RuleMatrix~\cite{Ming:2019:RuleMatrix}. RuleMatrix was applied to support the analysis of surrogate logic rules of Artificial Neural Networks and Support Vector Machine models. Local explanations are taken by analyzing the employed rules, observing the instance features values coupled with rules predicates and properties. Another interactive system closely related to our method is the iForest~\cite{Zhao:2019:iForest}, combining techniques for Random Forest models local explanations. The iForest system focuses on binary classification problems, and for each instance, it allows the exploration of decision paths from Decision Trees using multidimensional projection techniques. A summarized decision path is built and displayed as a node-link diagram by selecting decision paths of interest (circles in the projection).

As discussed before, node-link diagrams are prone to present scalability issues. Although iForest reduces the associate issues by summarizing similar decision paths, it fails to present the overall picture of Random Forest classification models' voting committees. Our approach shows the voting committee by displaying all rules (decision paths) used by a model when classifying a particular instance, allowing insights into the feature space and class association by ordering rules and features in different ways. Also, our approach can be applied to multi-class problems, not only binary classifications, and, as iForest, it supports counterfactual analysis~\cite{Gomez:2020:Vice, Liao:2020:Questioning}  by displaying the rules that, with the smallest changes, may cause the instance under analysis to switch its final classification.

\section{ExMatrix}

In this section, we present \textit{Explainable Matrix (ExMatrix)}, a visualization method to support Random Forest global and local interpretability.

\subsection{Overview}

To create a classifier, classification techniques take a labelled dataset ${X = \{x_{1},...,x_{N}\}}$ with $N$ instances and their classes ${Y=\{y_{1},...,y_{N}\}}$, where ${y_n \in C = \{c_{1},...,c_{J~\geq~2}\}}$ and $x_{n}$ consists of a vector ${x_{n} = [x_{n}^{1},...,x_{n}^{M}]}$ with $M$ features ${F = \{f_{1},...,f_{M}\}}$ values, and build a mathematical model to compute a class $y_n$ when new instances ${x_n \notin X}$ are given as input. In this process, $X$ is usually split into two different sets, one $X_{train}$ to build the model and one $X_{test}$ to test it. The existing techniques have adopted many different strategies to build a classifier. The Random Forest (RF) is an ensemble approach that creates multiple Decision Tree (DT) models ${DT_{1},...,DT_{K}}$ of randomly selected subsets of features and/or training instances, and combines them to classify an instance using a voting strategy~\cite{Tan:2005:Introduction, Breiman:1984:Classification, Breiman:2001:RandomForest, Biau:2016:Arandom}. Therefore, a RF model is a collection of decision paths, belonging to different DTs, combined to classify an instance. 

\begin{figure*}[h]
    \centering
    \includegraphics[width=\linewidth]{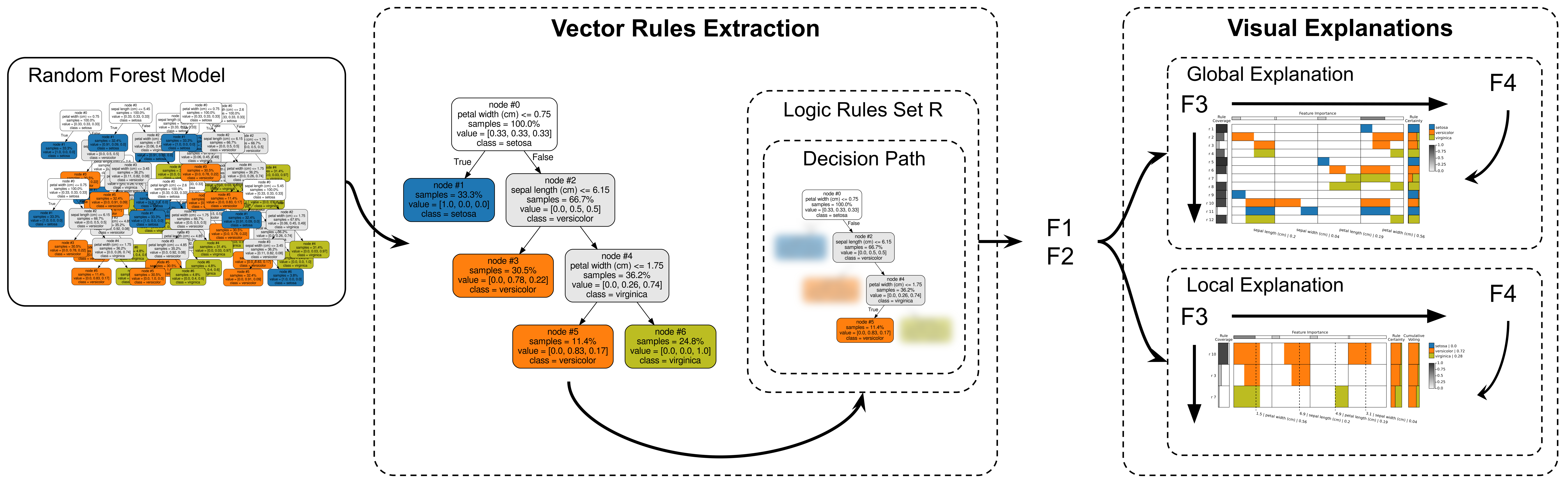}
    \caption{\textit{Explainable Matrix (ExMatrix)} overview. ExMatrix is composed of two main steps. In the first, decision paths of the RF model under analysis are converted into logic rules. Then, in the second, these rules are displayed using a matrix metaphor to support global and local explanations.}
    \label{fig:overview}
    \vspace{-0.4cm}
\end{figure*}

Aiming at supporting users to examine RF models and enable results audit, ExMatrix presents the decision paths extracted from DTs as logic rules using a matrix visual metaphor, supporting global and local explanations. ExMatrix arranges logic rules ${R = \{r_{1}, ..., r_{Z}\}}$ as rows, features ${F = \{f_{1},...,f_{M}\}}$ as columns, and rule predicates ${r_{z} = [ r_{z}^{1}, ..., r_{z}^{M} ]}$ as cells, inspired by similar user-friendly and powerful matrix-like solutions~\cite{Wu:2008:MatrixVisualization, Chen:2002:GAP, Chen:2004:MatrixVisualization}. \autoref{fig:overview} depicts our method overview, composed mainly of two steps. One involving the vector rules extraction, where all decision paths of each $DT_{k}$ in the RF model are converted into vectors, and a second one where these vectors are displayed using a matrix metaphor to support explanations. The next sections detail these steps, starting with the vector rule extraction process.

\subsection{Vector Rules Extraction}

As mentioned, ExMatrix first step involves the transformation of each decision path, the path from a DT root node to a leaf node, into a vector rule representing the features' intervals for which the decision path is true. The resulting vectors present dimensionality equal to the number of features $M$, with coordinates composed of pairs representing the features' minimum and maximum interval values. In more mathematical terms, this process transforms, for every tree $DT_{k}$, each decision path $p_{(o,d)}$ (from the root node $o$ to the leaf node $d$) into a disjoint logic rule (vector) $r_{z}$. Let $p_{(o,d)} = \{(f_{o} \bigotimes \theta_{o}),...,(f_{v} \bigotimes \theta_{v})\}$ denotes a decision path, where each node $i$ contains a logic test $\bigotimes \in \{``\leq",``>"\}$ bisecting the feature $f_{i}$ using a threshold $\theta_{i} \in \mathbb{R}$, and that the node $v$ is the parent of the leaf node $d$~\cite{Zhao:2019:iForest}. To convert $p_{(o,d)}$ into a vector rule ${r_{z} = [ r_{z}^{1}, ..., r_{z}^{M} ] }$, each element ${r_{z}^{m} = \{ \alpha_{z}^{m} , \beta_{z}^{m} \}}$ is computed representing the intervals covered by $p_{(o,d)}$ if and only if $f^{m} \in p_{(o,d)}$. Otherwise, $r_{z}^{m} = \varnothing$. Considering $f^{m} \in p_{(o,d)}$, the lower limit $\alpha_{z}^{m}$ is the maximum $\theta_i \in p_{(o,d)}$ for the feature $f^{m}$ and logic test $\bigotimes = ``>"$. If such combination does not exist in $p_{(o,d)}$, $\alpha_{z}^{m}$ is set to the minimum value of feature $f^m$ in $X$, that is
 
\vspace{-0.25cm}
\begin{equation*}
\alpha_{z}^{m} = \left\{ 
\begin{array}{ll} 
    \max( \theta_{i} | f_{i} = f^{m}, \bigotimes = ``>") & \text{if}~(f_i=f^{m} > \theta_i) \in p_{(o,d)}
    \\ 
    \min( x^{m} | x^{m} \in X ) & \text{Otherwise.}
\end{array}\right.
\label{eq:RealSetPredicateAlpha}
\end{equation*}

Similarly, the upper limit $\beta_{z}^{m}$ is the minimum $\theta_i \in p_{(o,d)}$ for the feature $f^{m}$ and logic test $\bigotimes = ``\leq"$. If such combination does not exist in $p_{(o,d)}$, $\beta_{z}^{m}$ is set to the maximum value of feature $f^m$ in $X$, that is
 
\vspace{-0.25cm}
\begin{equation*}
\beta_{z}^{m} = \left\{ 
\begin{array}{ll} 
    \min( \theta_{i} | f_{i} = f^{m}, \bigotimes = ``\leq" ) & \text{if}~(f_i=f^{m} \leq \theta_i) \in p_{(o,d)}
    \\ 
    \max( x^{m} | x^{m} \in X ) & \text{Otherwise.} 
\end{array}\right.
\label{eq:RealSetPredicateBeta}
\end{equation*} 

Beyond predicates, three other properties are extracted for each logic rule $r_{z}$, being certainty, class, and coverage. The rule certainty $r_{z}^{cert}$ is a vector of probabilities for each class $c_j \in C$, obtained from the decision path (leaf node value). The rule class $r_{z}^{class}$ is the $c_j \in C$ with the highest probability on the rule certainty $r_{z}^{cert}$. The rule coverage $r_{z}^{cov}$ is the number of instances in $X_{train}$ of class $r_{z}^{class}$ for which $r_{z}$ is valid divided by the total number of instance of $r_{z}^{class}$ in $X_{train}$. The vector rules extraction process results in a set of disjoint logic rules ${R = \{ r_{1}, ..., r_{Z} \}}$, where each rule $r_{z}$ classifies an instance $x_n$ belonging to class $r_{z}^{class}$ if its predicates ${r_{z} = [ r_{z}^{1}, ..., r_{z}^{M} ]}$ are all true for the feature values in $x_n$.

As an example of vector rule extraction, consider the zoomed DT in \autoref{fig:overview} from a RF for the Iris dataset~\cite{Fisher:1936:The}, with $150$ instances in three classes ${C = \{setosa, \; versicolor, \; virginica\}}$ and $4$ features $F = \{sepal\;length\;, \;sepal\;width\;, \;petal\;length\;, \;petal\;width\}$. From this tree, the decision path $p_{(\#0,\#5)}$ is transformed into the vector rule ${r_{3} = [ \{ 6.15 , 7.9 \}, \varnothing, \varnothing, \{ 0.75, 1.75 \} ]}$ with $r_{3}^{class} = versicolor$, since rule certainty equals to ${r_{3}^{cert} = [ 0.0, 0.83, 0.17 ]}$ (leaf node $\#5$ value), indicating that $r_{3}$ is valid for $83\%$ of the $versicolor$ instances and $17\%$ of $virginica$ instances in $X_{train}$. The rule coverage $r_{3}^{cov} = 0.28$ as $r_{3}$ is valid for $10$ out of $35$ $versicolor$ instances in $X_{train}$.

\subsection{Visual Explanations}

Once the vector rules are extracted, they are used to create the matrix visual representations for global and local interpretation. To guide our design process we adopted the iForest design goals (G1 - G3)~\cite{Zhao:2019:iForest} and the RuleMatrix target questions (Q1 - Q4)~\cite{Ming:2019:RuleMatrix} summarized on \autoref{tab:GuidelinesExp}. These goals and questions consider classification model reasoning beyond performance measures (e.g., accuracy and error), focusing on the model internals. For global explanations, where the focus is an overview of a model, ExMatrix displays feature space ranges and class associations (\textbf{G1} and \textbf{Q1}), and how reliable these associations are (\textbf{Q2}). For local explanations, where the focus is the classification of a particular instance $x_{n}$, ExMatrix allows the analysis of $x_{n}$ values and features space ranges that resulted into the assigned class $y_{n}$ (\textbf{G2} and \textbf{Q3}), and the inspection of the changes in $x_{n}$ that may lead to a different classification (\textbf{G3} and \textbf{Q4}).

\begin{table}[h]
    \caption{ExMatrix design goals.}
    \label{tab:GuidelinesExp}
    \scriptsize
    \centering%
    \begin{tabular}{ p{4cm} p{4cm} } 
        \toprule
        \multicolumn{1}{c}{Global} & \multicolumn{1}{c}{Local} \\
        \midrule
        \textbf{G1} Reveal the relationships between features and predictions~\cite{Zhao:2019:iForest}. & \textbf{G2} Uncover the underlying working mechanisms~\cite{Zhao:2019:iForest}. \\
        \textbf{Q1} What knowledge has the model learned?~\cite{Ming:2019:RuleMatrix} & \textbf{G3} Provide case-based reasoning~\cite{Zhao:2019:iForest}. \\
        \textbf{Q2} How certain is the model for each piece of knowledge?~\cite{Ming:2019:RuleMatrix} & \textbf{Q3} What knowledge does the model utilize to make a prediction?~\cite{Ming:2019:RuleMatrix} \\
         & \textbf{Q4} When and where is the model likely to fail?~\cite{Ming:2019:RuleMatrix} \\
        \bottomrule 
    \end{tabular}
\end{table}

ExMatrix implements these goals using a set of four functions:

\begin{description}
\item[F1 -- Rules of Interest.] Function $R'= f_{rules}(R, \ldots)$ returns a subset of rules of interest $R' \subseteq R$. For global explanations $f_{rules}(R, ...)$ returns the entire vector rules set $R' = R$ or a subset $R' \subset R$ defined by the user, while for local explanations $f_{rules}(R, x_{n}, ...)$ returns a subset $R' \subset R$ related to a given instance $x_{n}$.
    
\item[F2 -- Features of Interest.] Function $F' = f_{features}(R', \ldots)$ returns features of interest $F' \subseteq F$ considering a set of rules of interest $R'$. For global explanations $f_{features}(R', ...)$ returns all features used by the RF model, whereas for local explanations $f_{features}(R', x_{n}, ...)$ returns the features used to classify a given instance $x_{n}$. 

\item[F3 -- Ordering.] Function $L' = f_{ordering}(L, criteria, \ldots)$ returns an ordered version $L'$ of a input set $L$ following a given criterion, where $L$ can be rules $R'$ or features $F'$. This is used for both global and local explanations aiming at revealing patterns, a key property in matrix-like visualizations~\cite{Wu:2008:MatrixVisualization, Chen:2002:GAP, Chen:2004:MatrixVisualization}, where rows and columns can be sorted in different ways, following, for instance, elements properties~\cite{Krause:2017:AWorkflow} or similarity measures~\cite{Choi:2010:ASurvey, Tzeng:2009:Selection, Behrisch:2018:MatrixReordering, Fujiwara:2020:Supporting}. 

\item[F4 -- Predicate Icon.] Function $f_{icon}(r_{z}^{m}, \ldots)$ returns a cell icon (visual element) for a predicate $r_{z}^{m}$ of the rule $r_{z}$ and feature $f_{m}$. For global and local explanations, a cell icon is a color-filled rectangular element, allowing our visual metaphor to display a substantial number of logic rules at once. This is an important aspect since matrix-like visualizations can display a massive number of rows and columns relying on such icons not requiring many pixels~\cite{Chen:2004:MatrixVisualization}.
\end{description}

\autoref{fig:overview} shows how these four functions are used in conjunction to build the visual representations for global and local interpretation. Functions \textbf{F1} and \textbf{F2} are used to select and map rules and features of interest. Function \textbf{F3} is used to change the rows and columns order to help in finding interesting patterns, and function \textbf{F4} is used to derive the predicate icon that can vary depending on the type of interpretation task (global or local). In the next section, we detail how these functions are used to build ExMatrix visual representations.

\subsubsection{Global Explanation (GE)}
\label{sec:globalexp}

Our first visual representation is an overview of RF models called \textit{Global Explanation (GE)}. To build this matrix, $R'= f_{rules}(R, \ldots)$ returns all logic rules $R$ or a subset $R' \subset R$ defined by the user, and $F' = f_{features}(R', \ldots)$ returns all features used by at least one rule $r_{z} \in R'$. As previously explained, matrix rows represent logic rules, columns features, and cells rules predicates (icons). Rows and columns can be ordered using different criteria ($L' = f_{ordering}(L, criteria, \ldots)$). The rows can be ordered by rules' coverage, certainty, class \& coverage, and class \& certainty, while columns can be ordered by feature importance, calculated using the Mean Decrease Impurity (MDI)~\cite{Breiman:2002:Manual}. 

For the ExMatrix GE visualization, the matrix cell icon representing the rule predicate $r_{z}^{m}$ consists of a rectangle ($f_{icon}(r_{z}^{m}, \ldots)$) colored according to the rule class $r_{z}^{class}$, positioned and sized inside the matrix cell proportional to the predicate limits $\{ \alpha_{z}^{m} , \beta_{z}^{m} \}$, where the left side of the matrix cell represents the value $\min( x^{m} | x^{m} \in X )$ and the right side $\max( x^{m} | x^{m} \in X )$ (goals \textbf{G1} and \textbf{Q1}). The cell background not covered by the predicate limits can be either white or be filled using a less saturated color. If no predicate is present, the matrix cell is left blank. 

\begin{figure}
    \centering
    \includegraphics[width=\columnwidth]{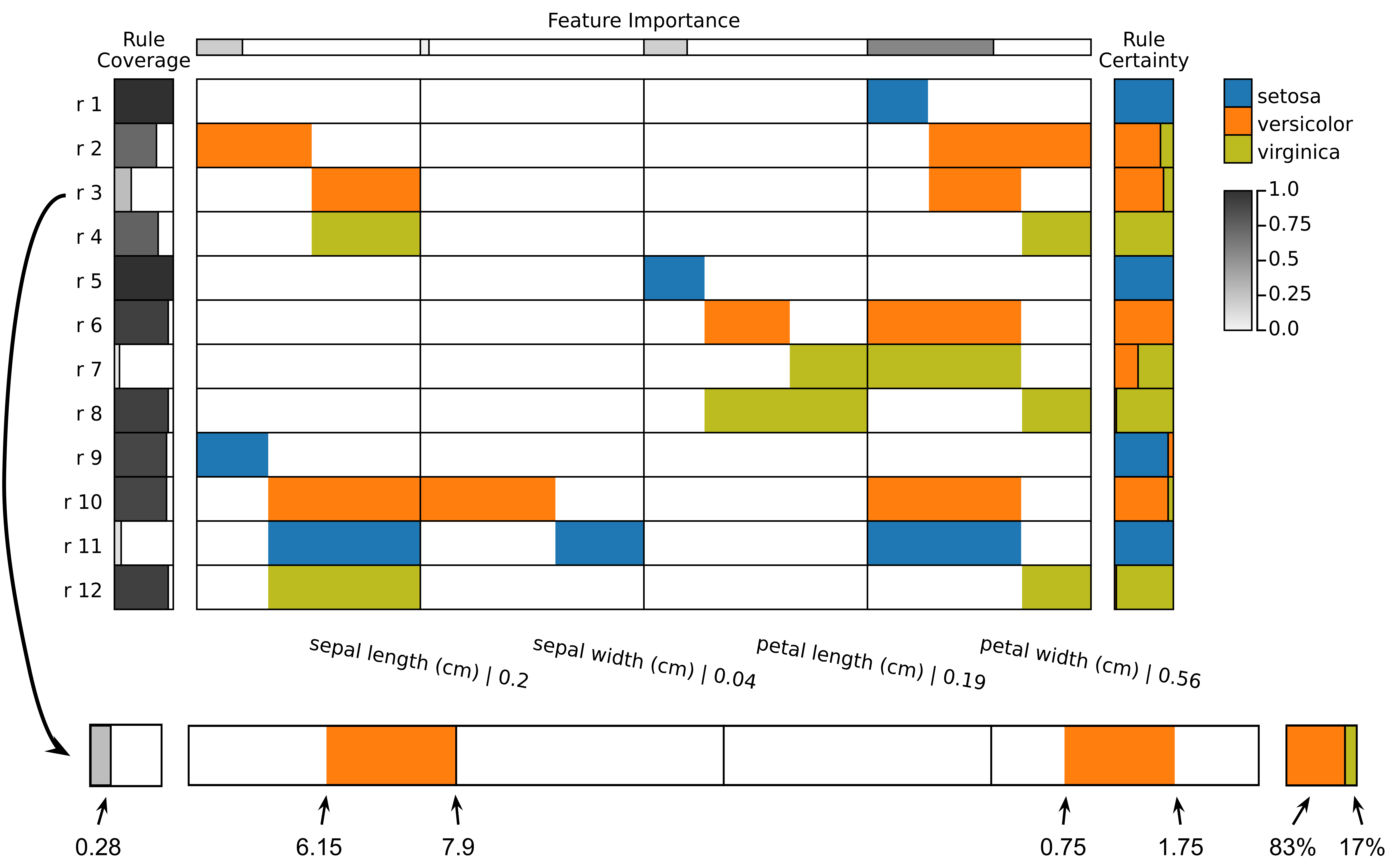}
    \caption{ExMatrix Global Explanation (GE) of a RF model for the Iris dataset containing $3$ trees with maximum depth equal to $3$.  Rows represent logic rules, columns features, and matrix cells the predicates. Additional rows and columns are also used to represent rule coverage and certainty. One matrix row is highlighted to exemplify how the rules' information is transformed into icons.}
    \label{fig:IrirRF3-3GlobalCoverxRaw-Values}
    \vspace{-0.4cm}
\end{figure}

Rules and features properties are also exhibited using additional rows and columns (goal \textbf{Q2}). The rule coverage $r_{z}^{cov}$ is shown using an extra column on the left side of the table with cells' color (grayscale) and fill proportional to the coverage. The rules certainty $r_{z}^{cert}$ is shown in an extra column in the right side of the table with cells split into colored rectangles with sizes proportional to the probability of the different classes. The feature importance is shown in an extra row on the top of the table with cells' color (grayscale) and fill proportional to the importance. Also, labels are added below the matrix, combining feature name and importance value.

\autoref{fig:IrirRF3-3GlobalCoverxRaw-Values} presents a ExMatrix GE visualization of a RF model for the Iris dataset with $3$ trees with maximum depth equals to $3$. In this example, the rows (rules) are ordered by extraction order, and the columns (features) follows the dataset order. The logic rule $r_{3} = [ \{ 6.15 , 7.9 \}, \varnothing, \varnothing, \{ 0.75, 1.75 \} ]$ extracted from the decision path $p_{(\#0,\#5)}$  (see \autoref{fig:overview}) is zoomed in. It is colored in orange since this is the color we assign to the $versicolor$ class and it classifies $83\%$ of the training instances as belonging to this class ($17\%$ belonging to $virginica$). Also, its coverage is  $r_{3}^{cov} = 0.28$.

\subsubsection{Local Explanation Showing the Used Rules (LE/UR)}
\label{sec:localexprules}

The second visual representation, called \textit{Local Explanation Showing the Used Rules (LE/UR)}, is a matrix to help in auditing the results of a RF model providing explanations for the classification of a given instance $x_{n}$. In this process, $R' = f_{rules}(R,x_{n})$ returns all logic rules used by the model to classify $x_{n}$ (goals \textbf{G2} and \textbf{Q3}). As in the ExMatrix GE visualization, $F'= f_{features}(R')$ returns all features used by logic rules $R'$,  $f_{icon}( r_{z}^{m}, X )$ returns a cell icon representing predicates limits, and $f_{ordering}(L, criteria)$ can order rules $R'$ by coverage, certainty, class \& coverage, and class \& certainty, and features $F'$ by importance. 

In addition to the coverage and certainty columns, in the ExMatrix LE/UR visualization, an extra column is added to represent the committee's cumulative voting. In this column, the cell at the $i^{th}$ row is split into colored rectangles with sizes proportional to the different classes' probability considering only the first $i$ rules. In this way, given a matrix order (e.g., based on the rule coverage), it is possible to see from what rule the committee reaches a decision that is not changed even if the remaining rules are used to classify $x_{n}$ (indicated by a black line). Notice that this column's last cell always represents the committee's final decision regardless of rule ordering.

\begin{figure}
    \centering
    \includegraphics[width=\columnwidth]{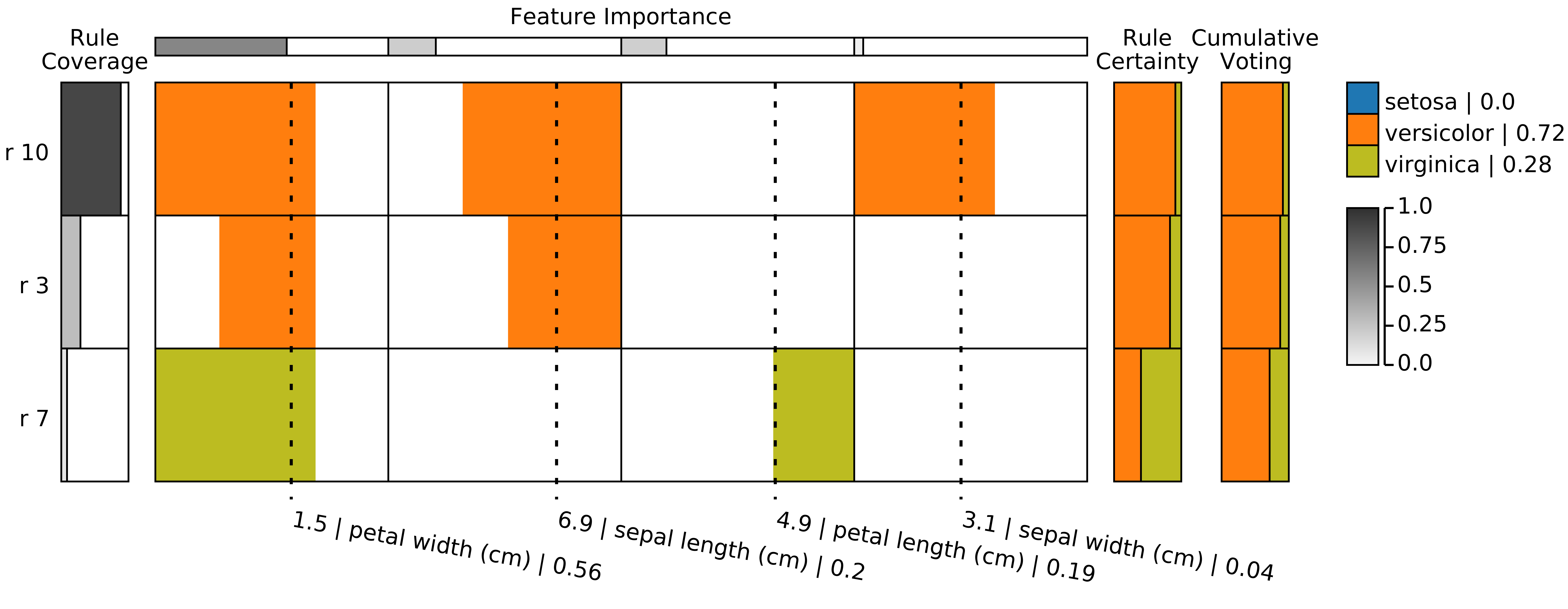}
    \caption{ExMatrix Local Explanation showing the Used Rules (LE/UR) visualization. Three rules are used by the RF committee to classify a given instance as belonging to the $versicolor$ class with $72\%$ of probability. The dashed line in each column indicates the features' values of the instance.}
    \label{fig:IrirRF3-3LocalUsed13CoverxImportance-Values}
    \vspace{-0.4cm}
\end{figure}

\autoref{fig:IrirRF3-3LocalUsed13CoverxImportance-Values} presents the ExMatrix LE/UR representation for instance $x_{13} = [6.9,3.1, 4.9, 1.5]$. We use the same RF model of \autoref{fig:IrirRF3-3GlobalCoverxRaw-Values} with $3$ trees, so the RF committee uses $3$ rules in the classification. The resulting matrix rows are ordered by rule coverage and columns by feature importance. The (optional) dashed line in each column indicates the values of the features of instance $x_{13}$.  According to the committee, the probability of $x_{13}$ to be $versicolor$ is $72\%$ and $28\%$ to be $virginica$. Most of the $virginica$ probability comes from the rule $r_{7}$, which holds the lowest coverage.

\subsubsection{Local Explanation Showing Smallest Changes (LE/SC)}

Our final matrix representation, called \textit{Local Explanation Showing Smallest Changes (LE/SC)}, is also designed to support results audit when classifying a given instance $x_{n}$. In this visualization, for each $DT_k$ in the RF model, we display the rule requiring the smallest change to make $DT_k$ to change the classification of $x_{n}$. Let $r_{z}$ be the rule extracted from $DT_{k}$ that is true when classifying $x_{n}$, in this process we seek for the rule $r_{e}$ from $DT_{k}$ with $r_{e}^{class} \neq r_{z}^{class}$ that presents the minimum summation of changes to the values of $x_n$ that makes $r_{e}$ true and $r_{z}$ false, that is, ${\Delta_{(r_{e},x_{n})} = \sum_{m = 1}^{M}( \Delta_{ (r_{e},x_{n}) }^{m} )}$, where 

\vspace{-0.25cm}
\begin{equation*}
\Delta_{ (r_{e},x_{n}) }^{m} = \left\{ 
\begin{array}{ll} 
    \frac{ \min( | \alpha_{e}^{m} - x_{n}^{m} |, | \beta_{e}^{m} - x_{n}^{m} | ) }{ | \max( x^{m} | x^{m} \in X_{train} ) - \min( x^{m} | x^{m} \in X_{train} ) | } & \text{if} \; x_{n}^{m} \notin \{ \alpha_{e}^{m} , \beta_{e}^{m} \} \\ 
0 & \text{Otherwise.} 
\end{array}\right.
\label{eq:DeltaChangeFeatureM}
\end{equation*}

Using this formulation, function $R'= f_{rules}(R,x_{n})$ returns the list of logic rules that can potentially change the classification process outcome requiring the lowest changes (goals \textbf{G3} and \textbf{Q4}), and function ${F' = f_{features}(R', x_{n})}$ returns the features used by the rules in $R'$. Beyond the ordering criteria for rules and features previously discussed, function $f_{ordering}(L, criteria)$ also allows ordering using the change summation ${\sum_{m = 1}^{M}( \Delta_{ (r_{e},x_{n}) }^{m} )}$. Finally, function $f_{icon}( r_{e}^{m}, x_{n} )$ returns a rectangle positioned and sized proportional to the change $\Delta_{ (r_{e},x_{n}) }^{m}$, with positive changes colored in green and negative in purple, with the cell matrix background filled using a less saturated color. If $\Delta_{ (r_{e},x_{n}) }^{m}=0$, the cell matrix is left blank. To help understand the class swapping, we add another column to the right of the table indicating the classification returned by the original rule $r_{z}$, showing the difference to the similar rule $r_{e}$ that cause the $DT_{k}$ to change prediction.

\begin{figure}
    \centering
    \includegraphics[width=\columnwidth]{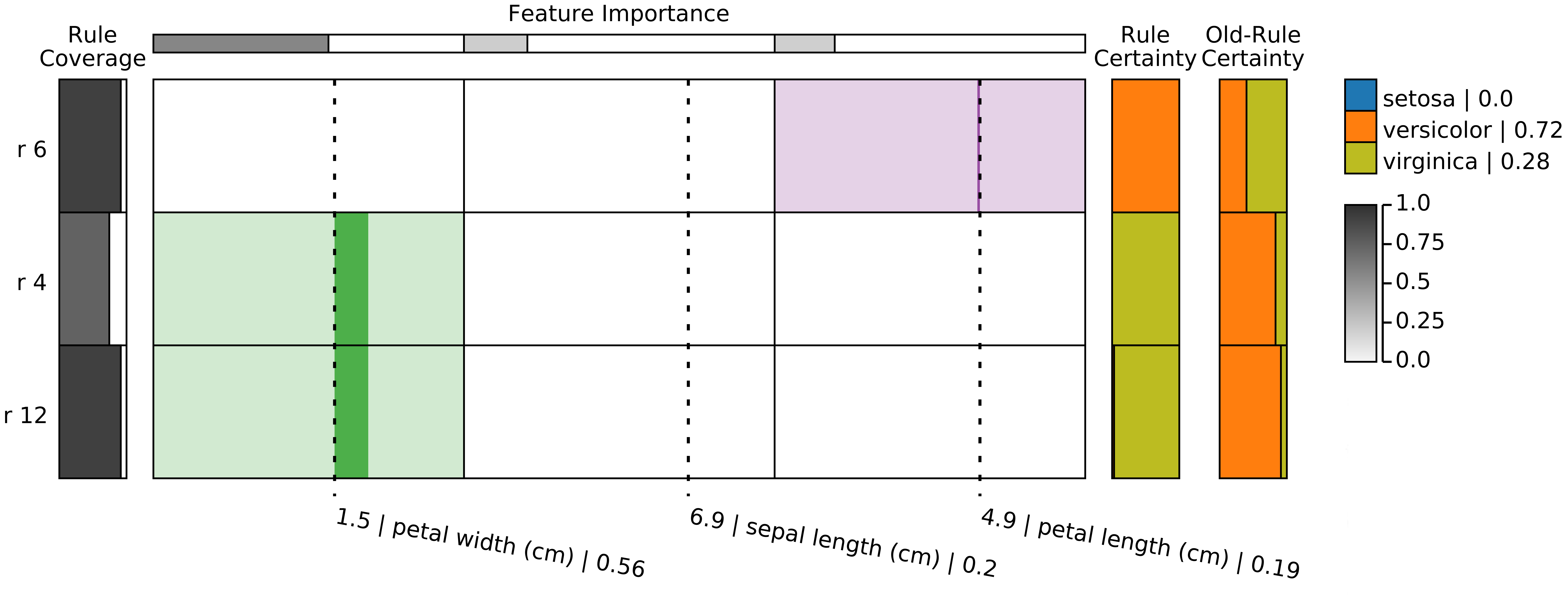}
    \caption{ExMatrix Local Explanation Showing Smallest Changes (LE/SC) visualization. Three rules with the smallest change to make the DTs to change class decisions are displayed. The rule in the first row presents the smallest change. Small perturbations may change the RF classification decision.}
    \label{fig:IrirRF3-3LocalClosest13DeltaChangexImportance-Values}
    \vspace{-0.4cm}
\end{figure}

\autoref{fig:IrirRF3-3LocalClosest13DeltaChangexImportance-Values} shows the ExMatrix LE/SC visualization for instance $x_{13} = [6.9,3.1, 4.9, 1.5]$ from the same RF model of \autoref{fig:IrirRF3-3GlobalCoverxRaw-Values}. Features $F'$ are ordered by importance and rules by change sum. The dashed lines represent the instance $x_{13}$ values. As an illustration, rule $r_{6}$ presents the smallest change in the feature ``petal length'' to replace a rule of majority class $virginica$ for a rule of class $versicolor$, potentially increasing the RF original outcome of 72\% for class $versicolor$ on instance $x_{13}$.

\section{Results and Evaluation}

In this section, we present and evaluate our method through a use-case~\footnote{\href{https://popolinneto.gitlab.io/exmatrix/papers/2020/ieeevast/usecase/}{https://popolinneto.gitlab.io/exmatrix/papers/2020/ieeevast/usecase/}} discussing the proposed features, two usage-scenarios~\footnote{\href{https://popolinneto.gitlab.io/exmatrix/papers/2020/ieeevast/usagescenarioi/}{https://popolinneto.gitlab.io/exmatrix/papers/2020/ieeevast/usagescenarioi/}}\footnote{\href{https://popolinneto.gitlab.io/exmatrix/papers/2020/ieeevast/usagescenarioii/}{https://popolinneto.gitlab.io/exmatrix/papers/2020/ieeevast/usagescenarioii/}} showing ExMatrix being used to explore RF models, finishing with a formal user test. All datasets employed in this section were downloaded from the \textit{UCI Machine Learning Repository}~\cite{Dua:2017}, and the ExMatrix implementation is publicly available as a Python package at \url{https://pypi.org/project/exmatrix/} to be used in association with the most popular machine learning packages.

\subsection{Use Case: Breast Cancer Diagnostic}
\label{subsec:usecase}

\begin{figure*}[ht]
    \centering
    \subfigure[ExMatrix GE visualization.]{\includegraphics[width=.76\linewidth]{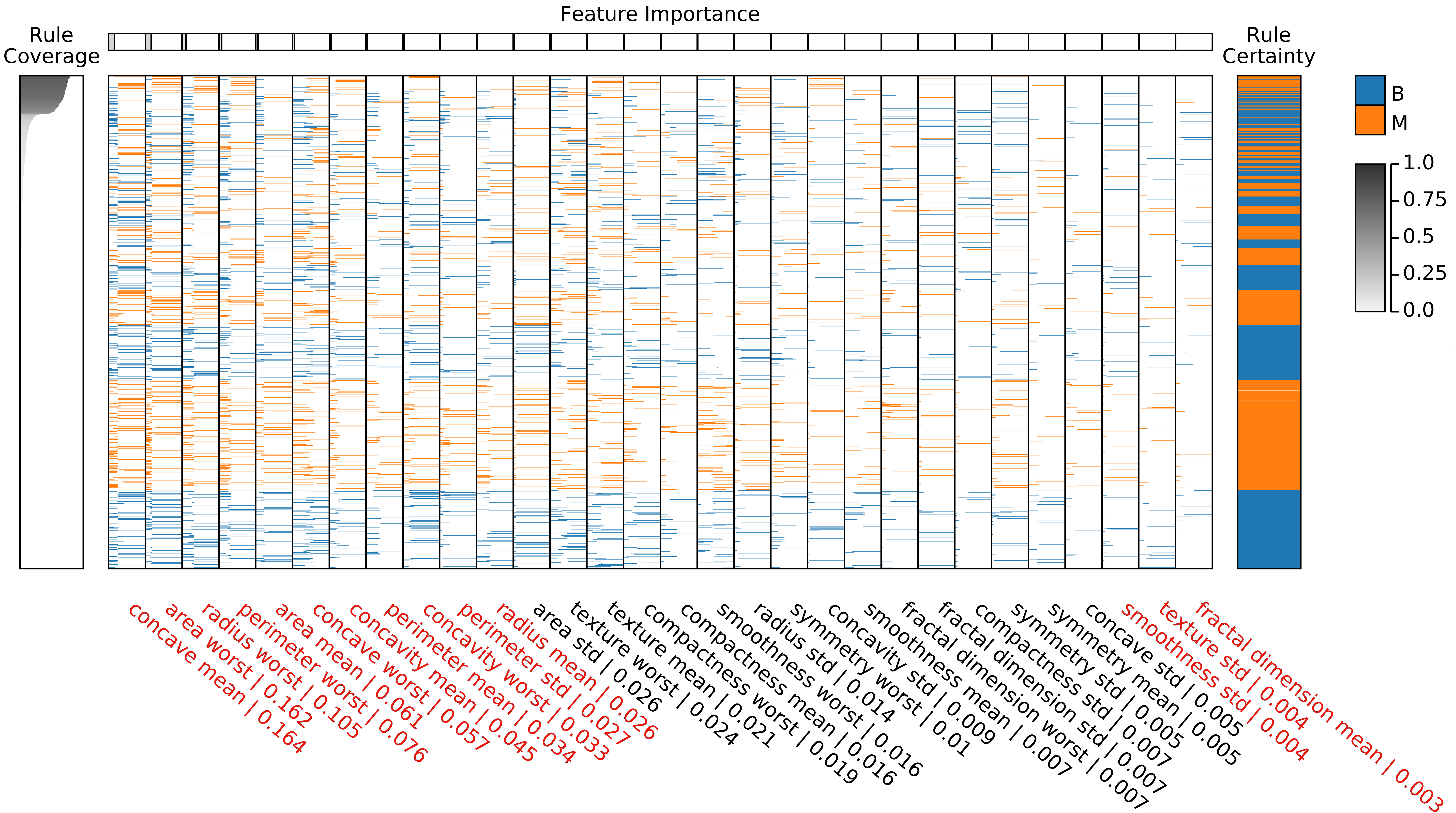}\label{fig:BreastCancerGE}}
    \subfigure[ExMatrix GE representation with filtered rules (only high-coverage rules).]{\includegraphics[width=.76\linewidth]{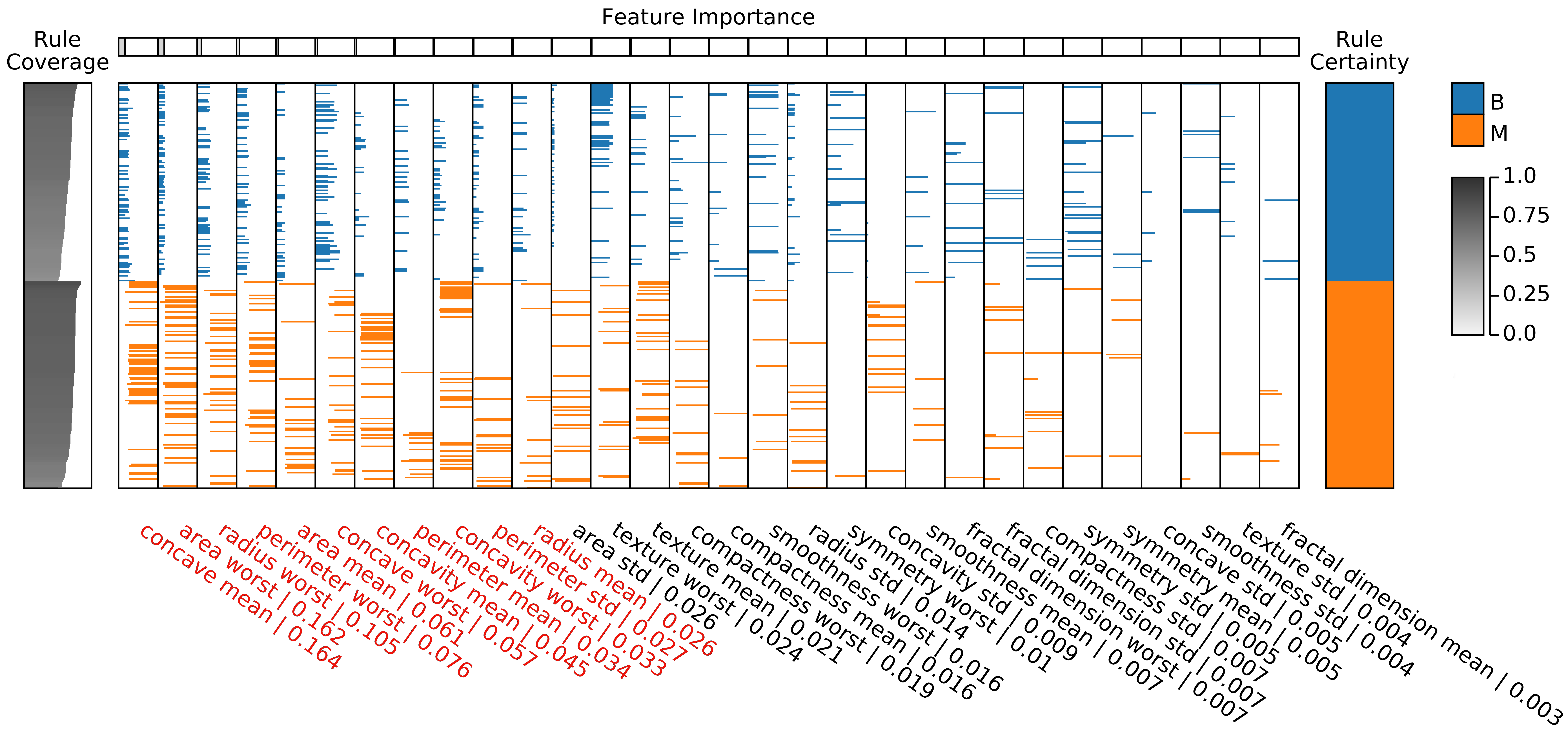}\label{fig:BreastCancerGE-F}}
     \vspace{-0.2cm}
    \caption{ExMatrix GE representations of the WDBC RF model. In (a), giving the ordering scheme by rule coverage and feature importance, some patterns emerge in terms of predicates ranges. In (b) the low-coverage rules are filtered-out to help focusing the analysis on what is important. Low feature values appear to be more related to class B whereas higher values to class M for the most important features.}
    \label{fig:testing}
     \vspace{-0.4cm}
\end{figure*}

In this use case, we utilize the \textit{Wisconsin Breast Cancer Diagnostic (WBCD)} dataset to discuss how to use ExMatrix global and local explanations to analyze RF models. The WDBC dataset contains samples of breast mass cells of $N = 569$ patients, $357$ classified as benign (B) and $212$ as malignant (M), with $M = 30$ features (cells properties). The RF model used as example was created randomly selecting $70\%$ of the instances for training and $30\%$ for testing and setting the number of DTs to $K = 128$, not limiting their depths. The result is a model with $3,278$ logic rules, $25.6$ rules per DT, and an accuracy of $99\%$.

An overview of this model is presented in Fig.~\autoref{fig:BreastCancerGE} using the ExMatrix GE representation (see \autoref{sec:globalexp}). In this visualization, rules are ordered by coverage and features by importance. Using this ordering scheme, it is possible to see that ``concave mean'', ``area worst'', and ``radius worst'' are the three most important features, whereas ``smoothness std'', ``texture std'', and ``fractal dimension mean'' are less important, and that the RF model used all $30$  features. Also, taking only the high coverage rules and features with more importance (``concave mean' to ``radius mean''), some patterns in terms of predicate ranges emerge. To help verify these patterns, low-coverage rules can be filtered out, resulting in a new visualization containing only high-coverage rules. Fig.~\autoref{fig:BreastCancerGE-F} presents the resulting filtered visualization with rules ordered by class \& coverage facilitating the comparison between the two dataset classes. In this new visualization, it is apparent that low feature values appear to be related to class B whereas higher values to class M (goals \textbf{G1}, \textbf{Q1}, and \textbf{Q2}). In this example, filtering aids in focusing on what is important regarding the overall model behavior, removing unimportant information and reducing cluttering, relying on the so-called Schneiderman's visualization mantra~\cite{Shneiderman:1996:Theeyes}.

\begin{figure*}[ht]
    \centering
    \subfigure[ExMatrix LE/UR for instance $x_{29}$, showing the used rules on the classification process.]{\includegraphics[width=.8\linewidth]{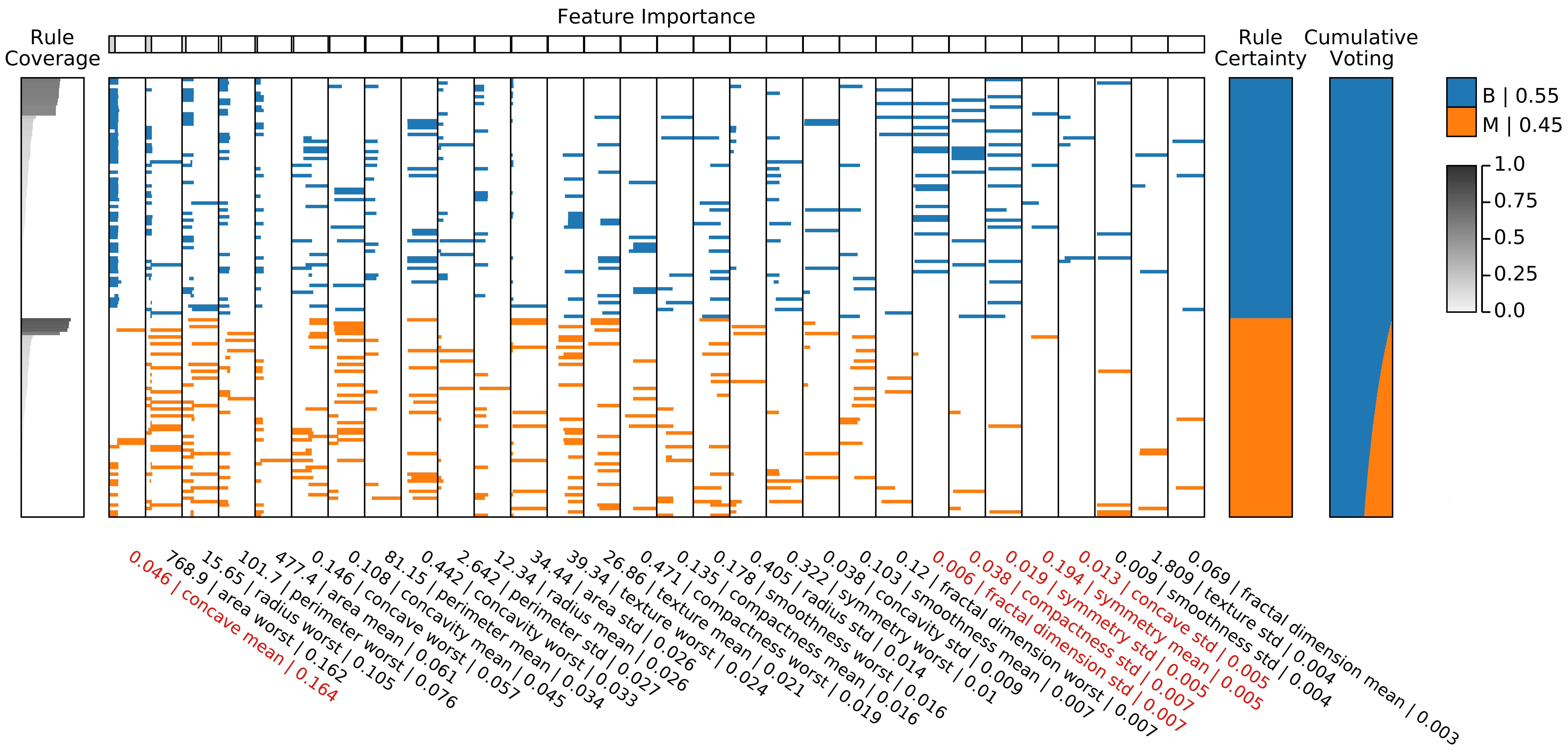}\label{fig:BreastCancerLEUR}}
    \subfigure[ExMatrix LE/SC for instance $x_{29}$, presenting changes in the instance feature values to make the DTs to change class prediction.]{\includegraphics[width=.8\linewidth]{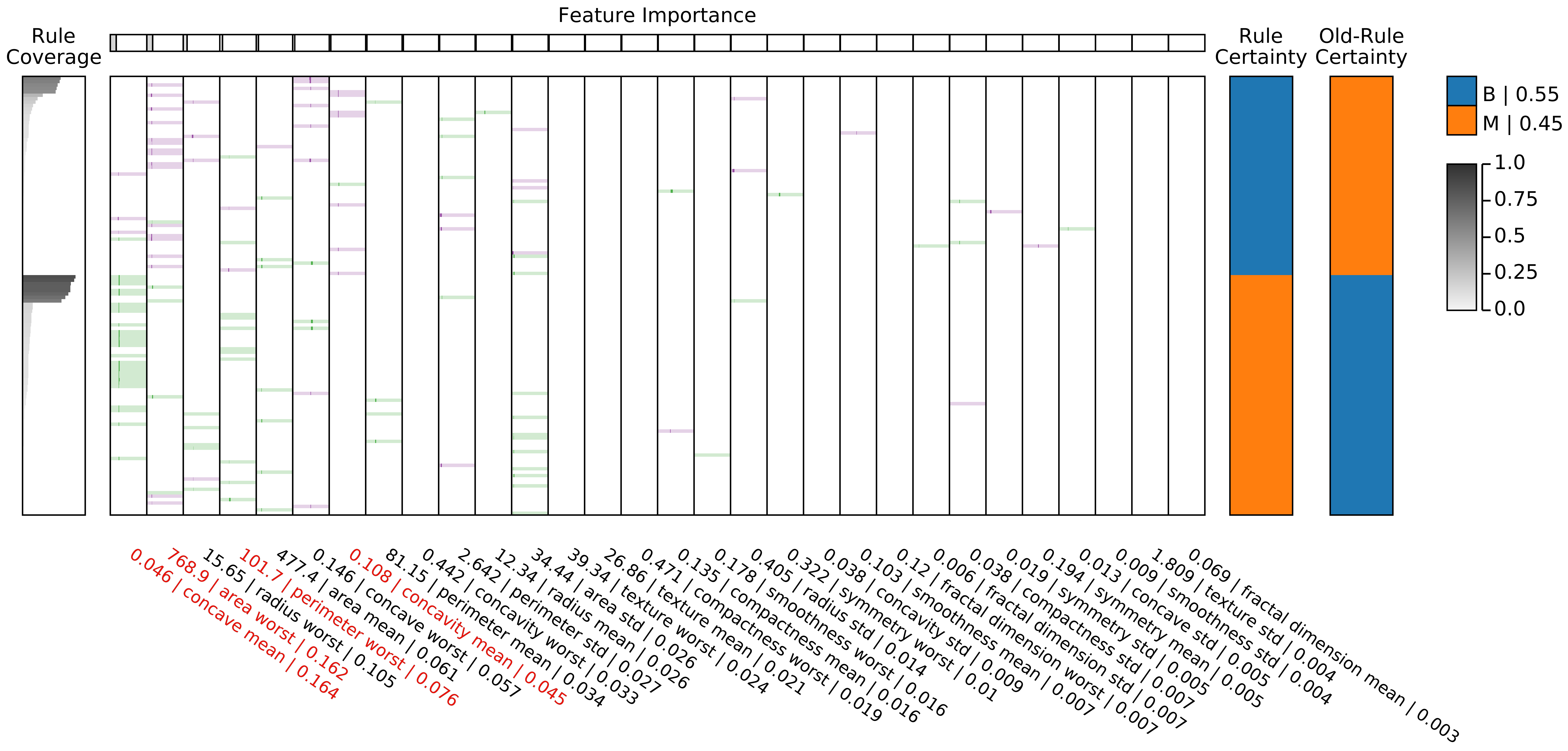}\label{fig:BreastCancerLESC}}
    \caption{ExMatrix local explanations of the WDBC RF model. Two different visualizations are displayed, one showing the rules employed in the classification of a target instance (a), and one presenting the smallest changes to make the trees of the model to change the prediction of that instance (b). In both cases, the target instance is the only misclassified instance.}
    \label{fig:testing}
     \vspace{-0.4cm}
\end{figure*}

The error rate of $1\%$ in this model is due to the misclassification of only one instance of the test set. Instance $x_{29}$ was wrongly classified as class B with a probability of $55\%$. Fig.~\autoref{fig:BreastCancerLEUR} shows the ExMatrix LE/UR representation (see \autoref{sec:localexprules}) using $x_{29}$ as target instance. In this visualization, the matrix is ordered by class \& coverage to focus on the difference between classes, and some interesting patterns are visible. For instance, predicate ranges of both classes B and M overlap for most features, except for ``fractal dimension std'' and ``concave std''. Also, these two features, along with ``symmetry std'', ``concave mean'', ``compactness std'', and ``symmetry mean'' are more related to class B (blue) since rules of such class heavily use them and sparsely used by rules of class M (orange) showing what is actively used by the model to make the prediction (goals \textbf{G2} and \textbf{Q3}). Besides, analyzing ExMatrix LE/SC visualization on Fig.~\autoref{fig:BreastCancerLESC}, one can notice that positive changes on features ``concave mean'' and ``perimeter worst'' may tie or alter the prediction of $x_{29}$ to class M since many green cells can be observed in the respective columns for rules of class M, while negative changes on ``area worst'' and ``concavity means'' increases its classification as class B since many purple cells can be observed in the respective columns for rules of class B (goals \textbf{G3} and \textbf{Q4}).

\subsection{Usage Scenario I: German Credit Bank}

As a first hypothetical usage scenario, we describe a bank manager Sylvia incorporating ExMatrix in her data analytics pipeline. To speed up the evaluation of loan applications, she sends her dataset of years of experience to a data science team and asks for a classification system to aid in the decision-making process. Such dataset contains $1,000$ instances (customers profiles) and $9$ features (customers information), with $700$ customers presenting rejected applications and $300$ accepted (here we use a pre-processed~\cite{Zhao:2019:iForest} version of German Credit Data from UCI). For the implementation of such a system, Sylvia has two main requirements: (1) the system must be precise in classifying loan applications, and; (2) the classification results must be interpretable so she can explain the outcome.

\begin{figure*}[!h]
    \centering
    \subfigure[ExMatrix GE showing rules filtered by coverage and certainty.]{\includegraphics[width=.485\linewidth]{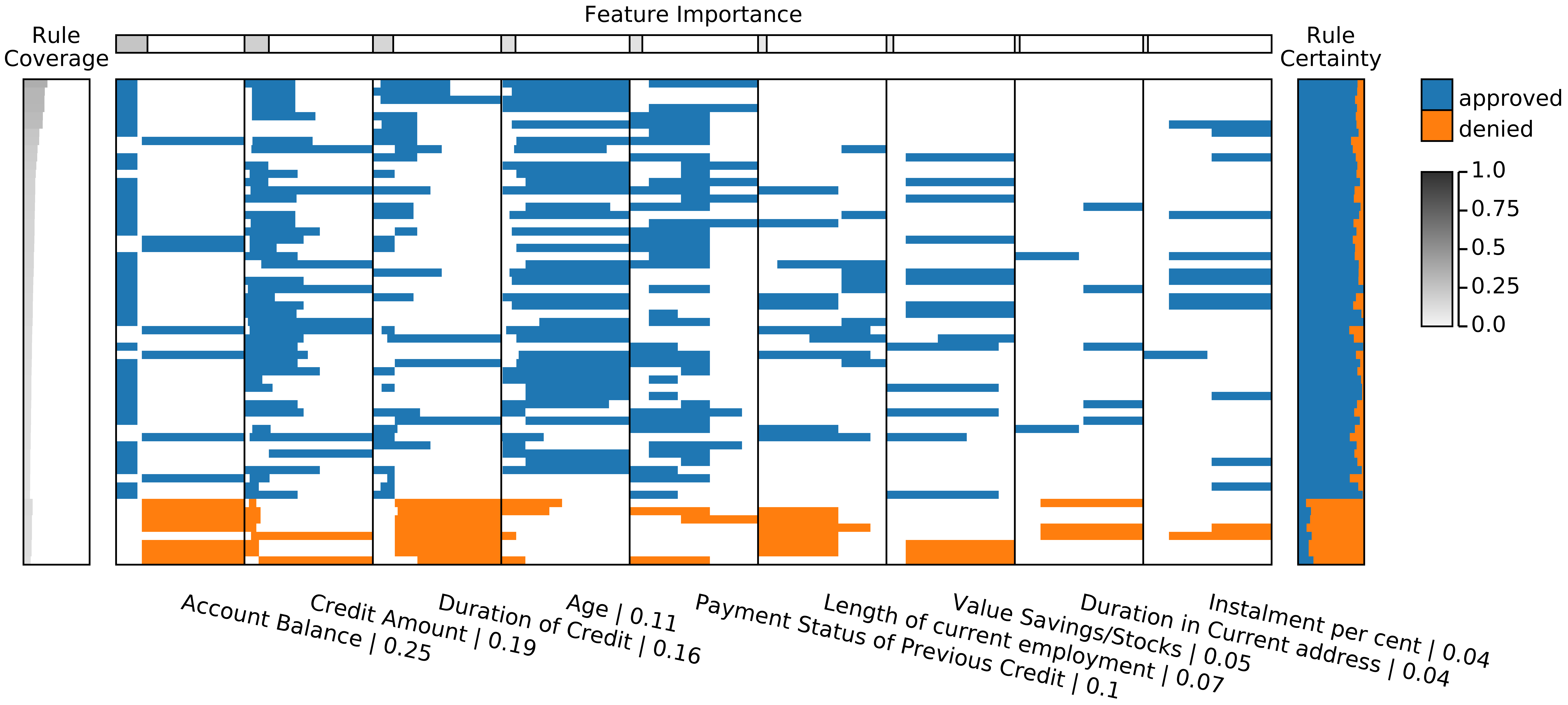}\label{fig:GermanCreditGE-F}}\quad
    \subfigure[ExMatrix LE/UR for instance $x_{154}$.]{\includegraphics[width=.485\linewidth]{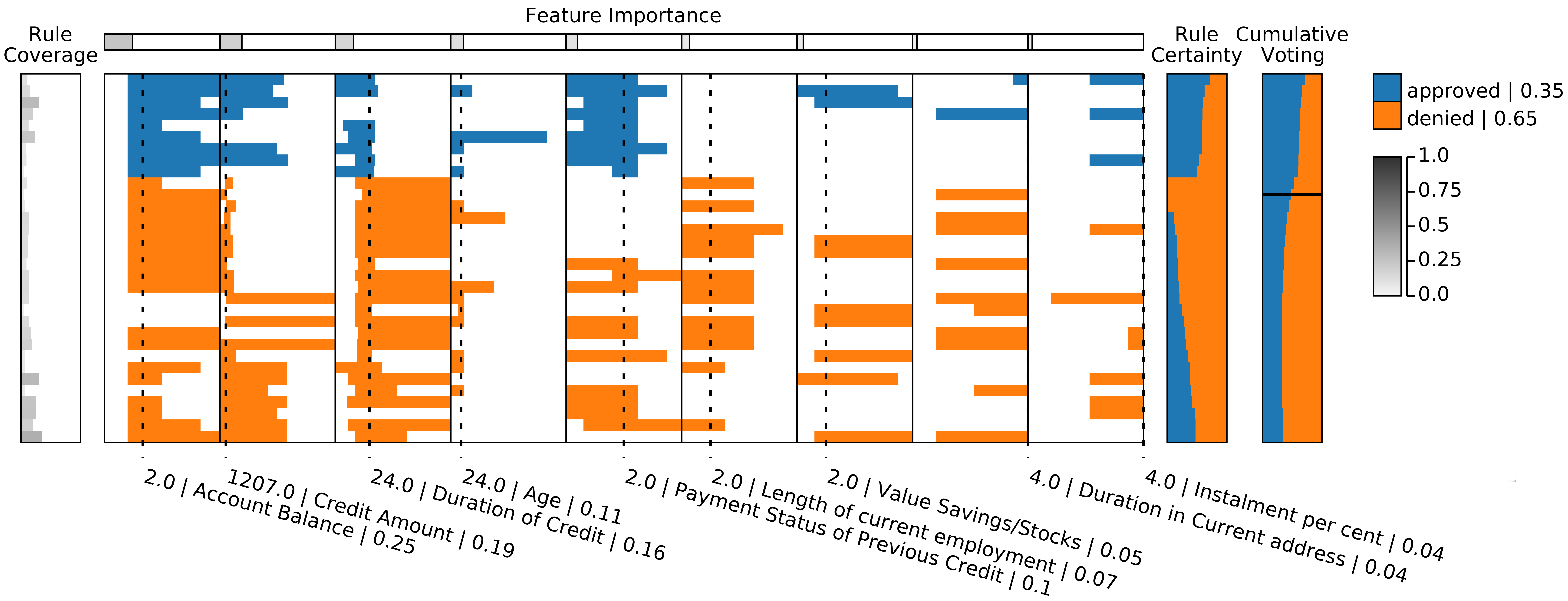}\label{fig:GermanCreditLEUR}}\\
    \subfigure[ExMatrix LE/SC for instance $x_{154}$.]{\includegraphics[width=.485\linewidth]{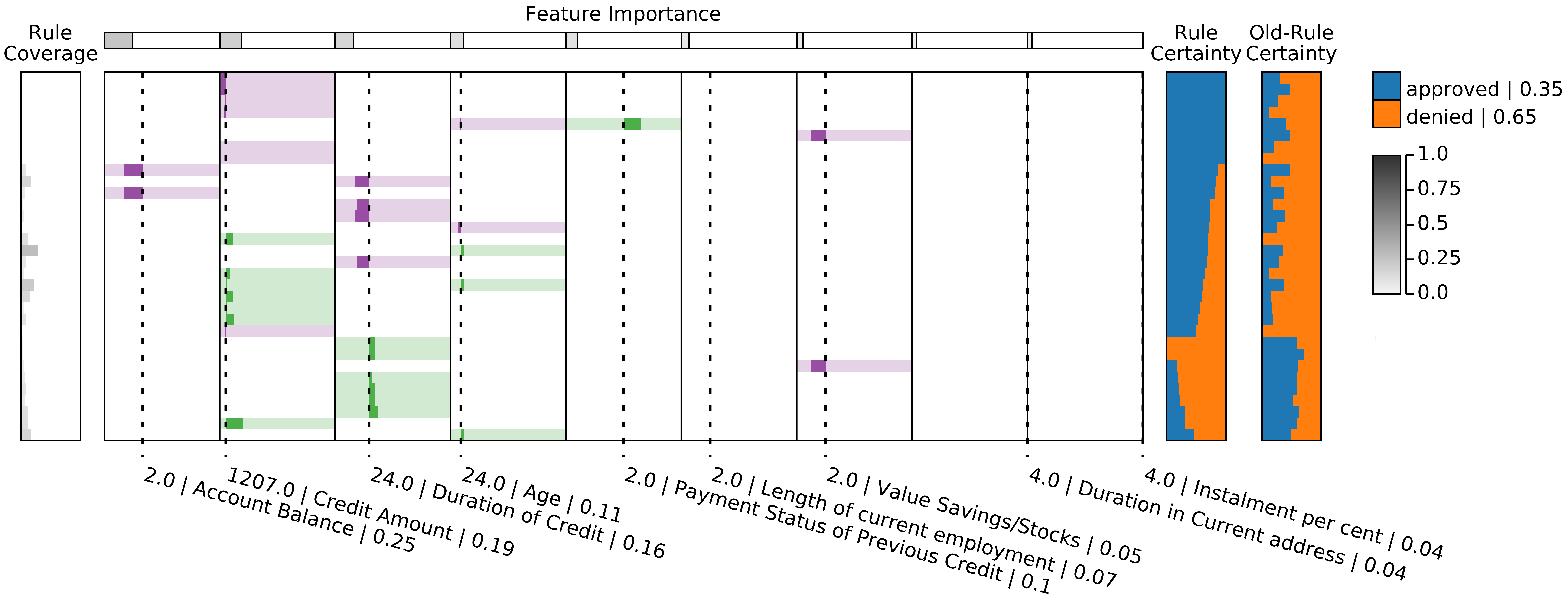}\label{fig:GermanCreditLESC}}\quad
    \subfigure[ExMatrix LE/UR modifying instance $x_{154}$, which changes RF's decision.]{\includegraphics[width=.485\linewidth]{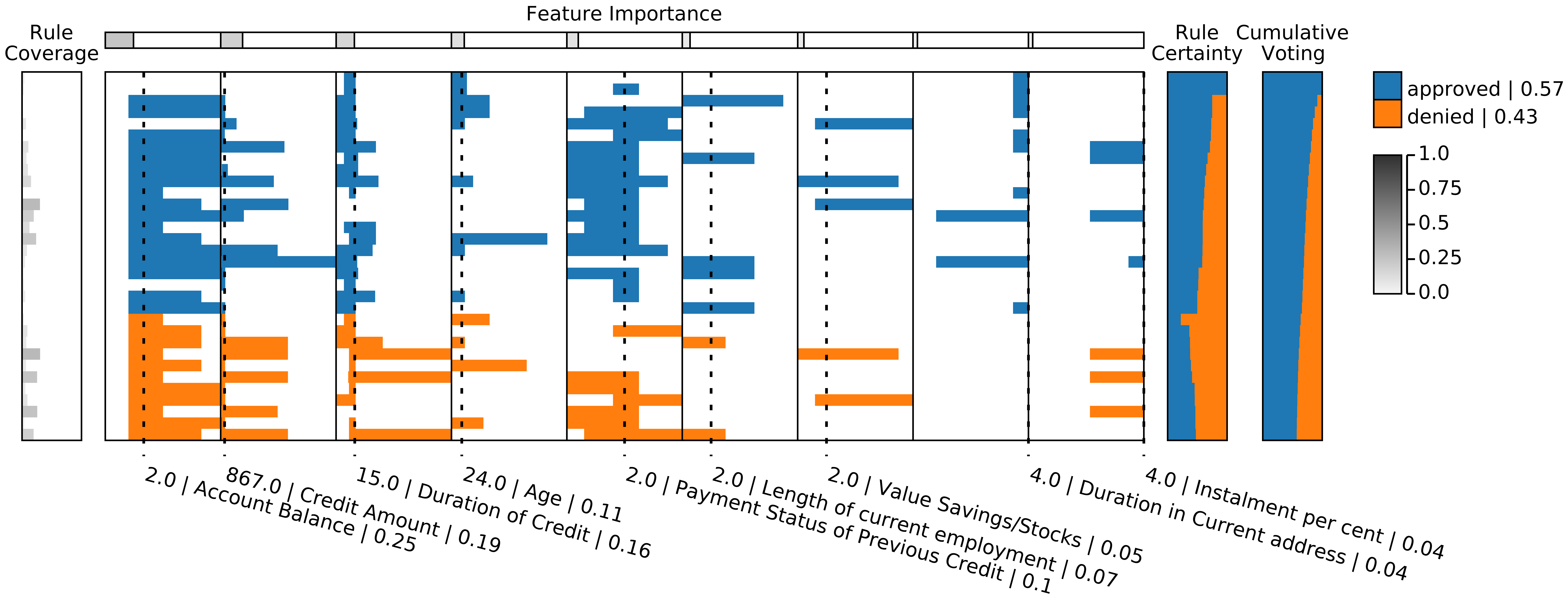}\label{fig:GermanCreditLEUR-N154}}
    \caption{ExMatrix explanations of a RF model for the German Credit Data UCI dataset. Based on the most generic knowledge learned by the RF model (rules with high coverage) (a), it is possible to conclude that applications requesting credit to be paid in longer periods tend to be rejected. Analyzing one sample (instance $x_{154}$) of rejected application (c), it is possible to infer that it is probably rejected due to the (applicant) short period working in the current job. However, lowering the requested amount as well as the number of instalments can change the RF's decision (d) and (e).}
    \label{fig::GermanCredit}
\end{figure*}

To fulfill the requirements, the data science team builds an RF model setting the number of DTs to $32$ with a maximum depth of $6$. The produced model's accuracy was $81\%$, resulting in $1,273$ logic rules, $38.7$ rules per DT. Using the ExMatrix GE representation (omitted due to space constraints, see supplemental material), she observes that the features ``Account Balance'', ``Credit Amount'', and ``Duration of Credit'' are the three most important, whereas ``Value Savings/Stocks'', ``Duration in Current address'', and ``Instalment percent'' are the three less. Also, by inspecting the most generic knowledge learned by the system (patterns formed by high-coverage rules) using a filtered representation of the ExMatrix GE visualization on Fig.~\autoref{fig:GermanCreditGE-F}, she notices that applications that request a credit to be paid in more extended periods (third column) tend to be rejected, matching her expectations. However, unexpectedly, customers without account (``Account Balance'': 1 - No account, 2 - No balance, 3 - Below $\$200$ , 4 - $\$200$ or above) have less chance to have their application rejected (first column), something she did not anticipate (goals \textbf{G1}, \textbf{Q1}, and \textbf{Q2}). Although confronting some of her expectations and bias, she trusts her data, and the classification accuracy seems convincing, so she decides to put the system in practice.

One day she receives a new customer interest in a loan. After filling the system with his data, unfortunately, the application got rejected by the classification system. Based on the new \textit{European General Data Protection Regulation}~\cite{Liu:2018:Interpretable, Carvalho:2019:Machine, Guidotti:2018:Survey} that requires explanations about decisions automatically made, the customer requests clarification. By inspecting the ExMatrix LE/UR visualization on Fig.~\autoref{fig:GermanCreditLEUR}, she notices, besides the denied probability of $65\%$, that even if all ``approved'' rules (blue) are used, very few high-certainty ``denied'' rules (orange) define the final decision of the model (see the Cumulative Voting and Rule Certainty columns), indicating that those rules, and the related logic statements, have a strong influence in the loan rejection. Also, she sees that the feature ``Length of current employment'' is the most directly related to the denied outcome since it is used only by rules that result in rejection (goals \textbf{G2} and \textbf{Q3}). Using this information, she explains to the customer that since he is working for less than one year in the current job ($2$ as ``Length of current employment'' corresponds to less than $1$ year), the bank recommends denying the application. However, analyzing the ExMatrix LE/SC representation in Fig.~\autoref{fig:GermanCreditLEUR}, she realizes that negative changes in features ``Credit Amount'' and ``Duration of Credit'' may turn the outcome to approved (goals \textbf{G3} and \textbf{Q4}). Thereby, as an alternative, she suggests lowering the requested amount and the number of installments. Based on the observable differences to make the rules change class, she notices that upon reducing the credit application from $\$1,207$ to $\$867$ and the number of payments from $24$ to $15$, the system changes recommendation to ``approved''. Fig.~\autoref{fig:GermanCreditLEUR-N154} presents the ExMatrix LE/UR visualization if such suggested values are used, changing the final classification.

\subsection{Usage Scenario II: Contraceptive Method}

This last usage scenario presents Christine, a public health policy manager who wants to create a contraceptive campaign to advertise a new, safer drug for long term use. To investigate married wives' preferences, Christine's data science team creates a prediction model using a data set with information about contraceptive usage choices her office collected past year (here we use the Contraceptive Method Choice dataset from UCI). The dataset contains $1,473$ samples (married wives profiles) with $9$ features, where each instance belongs to one of the classes ``No-use'', ``Long-term'', and ``Short-term'', regarding the contraceptive usage method, with $42.7\%$ of the instances belonging to class No-use,  $22.6\%$ to Long-term, and $34.7\%$ to Short-term. 

Since interpretability is mandatory in this scenario, allowing the results to be used in practice, the data science team creates an RF model and employs ExMatrix to support analysis. To create the model, the team set the number of DTs to $32$ and maximum depth to $6$, resulting in $1,383$ logic rules, $43.2$ rules per DT. The RF model accuracy is $63\%$, and, although not ideal for individual classifications, can be used to understand general knowledge learned by the model from the dataset. 

\begin{figure}[!h]
    \centering
    \includegraphics[width=\columnwidth]{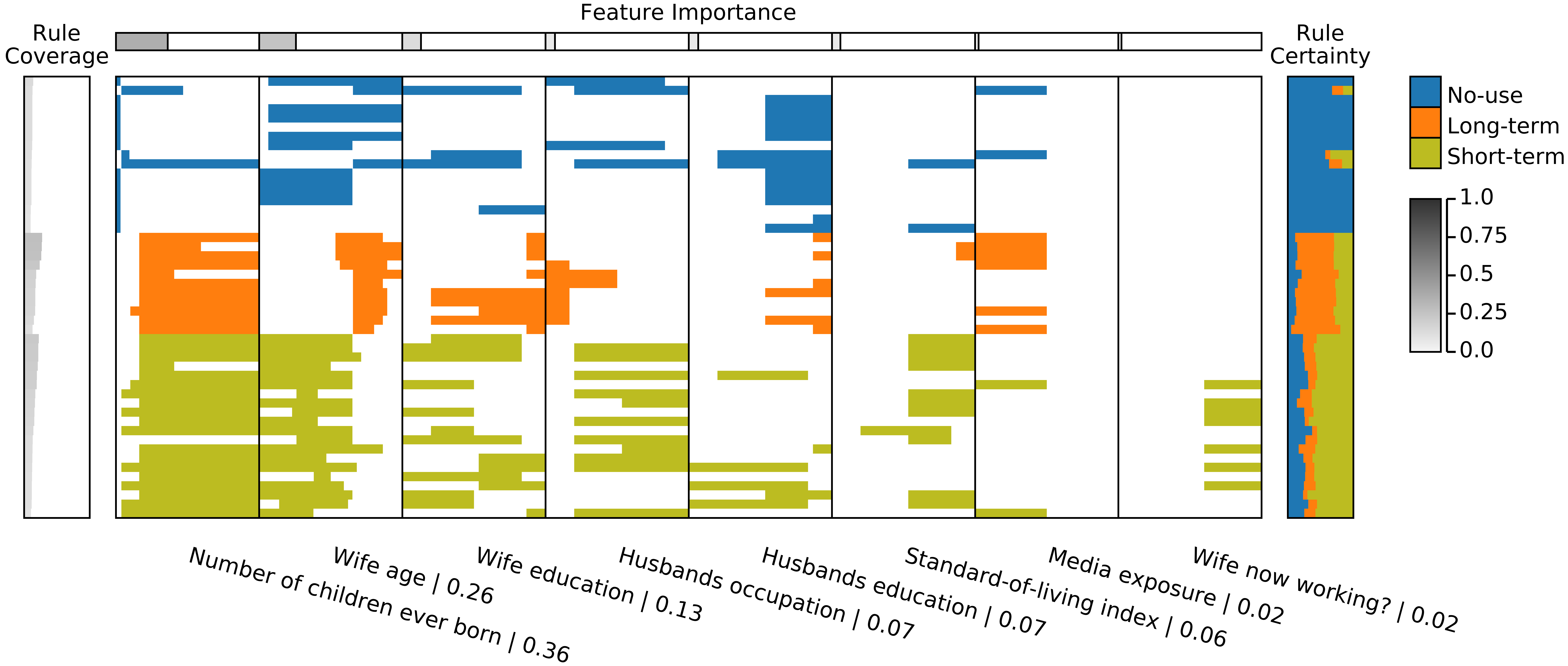}
    \caption{ExMatrix GE representation (rules filtered by coverage and certainty) of the RF model for the Contraceptive Method Choice UCI dataset. Based on high-coverage high-certainty rules, some interesting patterns can be observed. For instance, on contraceptive method usage, older women tend to use long-term contraceptive methods.}
    \label{fig:ContraceptiveMethodChoiceGE-F}
     \vspace{-0.4cm}
\end{figure}

By inspecting the ExMatrix GE representation of the model (omitted due to space constraints, see supplemental material), she readily understands that the features ``Number of children ever born'', ``Wife age'', and ``Wife education'' are the three most relevant for defining the contraceptive method class, while ``Media exposure'', ``Wife now working?'', and ``Wife religion'' are the three less. Also, further exploring a filtered version of the ExMatrix GE representation on \autoref{fig:ContraceptiveMethodChoiceGE-F}, to focus only on high-coverage and high-certainty rules ordered by class, she notices some interesting patterns regarding features space ranges and classes. For instance, lower values for the feature ``Number of children ever born'' (first column) are more related to class No-use and rarely related to class Long-term. For contraceptive method usage, higher values for the feature ``Wife age'' (second column) are related to class Long-term, while average and lower values are more related to class Short-term. Also, higher values for ``Wife education'' (third column) are more related to class Long-term (goals \textbf{G1}, \textbf{Q1}, and \textbf{Q2}). Based on these observations, and given the modest budget she received for the campaign, Christine decides to focus on the group of older and highly educated wives with at least one child to target the campaign's first phase.

\subsection{User Study}

\begin{table*}[h]
    \caption{User study questions.}
    \label{tab:AnalysisTestQuestions}
    \scriptsize
    \centering
    \begin{tabular}{ p{13cm} p{1.75cm} p{1.25cm} } 
        \toprule
        
        Question & Goals & Visualization \\
        \midrule
                
        \textbf{Qst 1} - About features space ranges and class ASSOCIATIONS. Considering rules with HIGH COVERAGE, and features with HIGH IMPORTANCE, select your answer: (three options of associations) & G1, Q1, and Q2 & Fig.~\autoref{fig:BreastCancerGE} \\
        
        \textbf{Qst 2} - Instance 29 is classified as Class A with a probability of 55\%, against 45\% for Class B. What feature is more related to Class A and less related to Class B? (four options of features) & G2 and Q3 & Fig.~\autoref{fig:BreastCancerLEUR}  \\

        \textbf{Qst 3} - Select the pair of features where DELTA CHANGES on instance 29 will potentially INCREASE Class A probability, and by that may SUPPORT its classification as Class A. (four options of features pairs) & G3 and Q4 & Fig.~\autoref{fig:BreastCancerLESC}  \\
        
        \textbf{Qst 4} - Select the pair of features where DELTA CHANGES on instance 29 will potentially INCREASE Class B probability, and by that may ALTER its classification as Class A. (four options of features pairs) & G3 and Q4 & Fig.~\autoref{fig:BreastCancerLESC} \\
        
        \bottomrule 
    \end{tabular}
     \vspace{-0.4cm}
\end{table*}

To evaluate the ExMatrix method, we performed a user study to assess the proposed visual representations for global and local explanations.  In this study, we asked four different questions based on the ExMatrix visualizations created for the use-case presented in \autoref{subsec:usecase}, focusing on evaluating the goals presented in \autoref{tab:GuidelinesExp}.

The study started with video tutorials about RF basic concepts and how to use ExMatrix to analyze RF models and classification results through the proposed explanations. A total of $13$ users participated, $69.2\%$ male and $30.8\%$ female, aged between $24$ to $36$, all with a background in machine learning. The participants were asked to analyze the explanations of Fig.~\autoref{fig:BreastCancerGE}, Fig.~\autoref{fig:BreastCancerLEUR}, and Fig.~\autoref{fig:BreastCancerLESC}, where each analysis was followed by different question(s) (see \autoref{tab:AnalysisTestQuestions}). On the visualizations, features names were replaced by ``Feature 1'' to ``Feature 30'' and classes names by ``Class A'' and ``Class B'', aiming at removing any influence of knowledge domain in the results, since our focus is to assess the visual metaphors.

Using the ExMatrix GE representation (Fig.~\autoref{fig:BreastCancerGE}), $76.9\%$ of the participants were able to identify patterns involving feature space ranges and classes, where, for high coverage rules, low features values are more related to class B, while features with large values are more related to class M (\textbf{Qst 1}). Using the ExMatrix LE/UR (Fig.~\autoref{fig:BreastCancerLEUR}), also $76.9\%$ of the participants were able to recognize that feature ``concave std'' is the most related to class B for instance $x_{29}$ classification outcome (\textbf{Qst 2}). Using the ExMatrix LE/SC (Fig.~\autoref{fig:BreastCancerLESC}), $61.5\%$ of the participants were able to identify that negative changes on instance $x_{29}$  features ``area worst'' and ``concavity mean'' values would better support the class B outcome (\textbf{Qst 3}), and $46.2\%$  were able to identify that positive changes on features ``concave mean'' and ``perimeter worst'' values may alter the outcome from class B to class M (\textbf{Qst 4}).

In general, the results were promising for the first two analyses, but the participants present worse results when interpreting the ExMatrix LE/SC visualization. This is not surprising since this representation requires a much better background about RF theory. The ExMatrix GE and LE/UR visualizations are more generic and involve much fewer concepts about how RF models work internally. In contrast, the ExMatrix LE/SC requires a good level of knowledge about ensembles models and how the voting system work when making a prediction. Although most of the users self-declared with a background in machine learning, only $30\%$ are RF experts. 

We also have asked subjective, open questions, and, in general, users gave positive feedbacks about ExMatrix explanations, where the visualizations were classified as visually pleasing and useful for understanding RF models.

\section{Discussion and Limitations}

%\subsection{Why NOT a Tree Structure Visual Metaphor?}

Although the natural choice to visualize a tree collection is to use tree structure metaphors, two main reasons make disjoint rules organized into tables a better option when analyzing DTs and especially RFs. First, using tree structure metaphors, the visual comparison of logic rules (decision paths) can be overwhelming since different paths from the root to the leaves define different orders of attributes, slowing down users when searching within a tree to answer classification questions~\cite{Freitas:2014:Comprehensible, Huysmans:2011:Anempirical}. An issue that is amplified in RFs, since multiple DTs are analyzed collectively. In contrast, in a matrix metaphor, the attributes are considered in the same order easing this process~\cite{Freitas:2014:Comprehensible, Huysmans:2011:Anempirical}. Second, given the constraints of usual DT inference methods (non-overlapping predicates with open intervals), features can be used multiple times in a single decision path resulting in multiple nodes (one per test) using the same feature. Consequently, if tree structures are employed, each feature's decision intervals need to be mentally composed by the user, and nodes using the same feature can be far away in the decision path. The decision intervals are explicit in the matrix representation and can be easily compared across multiples rules and trees. Therefore, although tree structure visual metaphors are the usual choice when hierarchical structures are the focus~\cite{Graham:2009:Asurvey, Schulz:2011:Thedesign}, on DTs and RFs, the decision paths are the object of analysis~\cite{Tan:2005:Introduction, Freitas:2014:Comprehensible, Huysmans:2011:Anempirical, Lima:2009:Domain} and transforming paths into disjoint rules organized into tables emphasize what is essential (see supplemental material).

%\subsection{Visual Scalability Issues}

Considering the above points, it is clear that scalability for RFs visualization is not just a choice of getting a visual metaphor that can represent millions of nodes, but getting a visual representation that is scalable and still properly supports essential analytical tasks (see Table~\ref{tab:AnalysisTestQuestions}). Something much more complex than merely visualizing a forest of trees. In this scenario, ExMatrix renders a promising solution, supporting the analysis of many more rules concomitantly than the existing state-of-the-art techniques. However, it is not a perfect solution. ExMatrix covers two different perspectives of RFs, conveying Global and Local information. In the Local visualization, scalability is not a problem since one rule is used per DT, so even for RFs with hundreds or even thousands of trees, ExMatrix scales well. However, for Global visualization, scalability can be an issue since the number of rules substantially grows with the number of trees. Although we can represent one rule per line of pixels, we are limited by the display resolution, and, even when the display space suffices, ExMatrix layouts can be cluttered and tricky to explore.

The solution we adopt to address scalability was to implement the so-called Schneiderman's visualization mantra~\cite{Shneiderman:1996:Theeyes}, allowing users to start with an overview of the model, getting details-on-demand by filtering rules to focus on specific sets of interest. Although users are free to select any subset of rules, considering that the goal of the Global visualization is to generate insights about the overall models' behavior, here we mainly explore filtering low-coverage rules since they are only valid for a few specific data instances (that is the coverage definition). Although simple, such a solution makes the analysis of entire models easier by removing unimportant information and reducing cluttering. Another potential solution is to make the rows' height proportional to coverage or certainty so that the rules with the lowest coverage or certainty are less prominent (visible) and could even be combined in less than one line of pixels. We have not tested this approach and left it as future work.

%\subsection{User Study Results}

Regarding the user study, although the results were satisfactory and within what we expect for the ExMatrix GE and LE/UR visualizations, the results for the ExMatrix LE/SC representation were sub-optimal, and the \textit{XAI Question Bank}~\cite{Liao:2020:Questioning} can help us to shed some light about the reasons. According to this bank, the GE addresses the leading question ``\textit{How (global)}'', whereas the LE/UR addresses the leading question ``\textit{Why}'', enabling to answer inquiries such as``\textit{What are the top rules/features it uses?}'' and ``\textit{Why/how is this instance given this prediction?}''.  However, the LE/SC involves three leading questions, ``\textit{What If}'', ``\textit{How to be that}'', and ``\textit{How to still be this}'', where the changes on instance feature values are presented supporting hypotheses (not answers), which shown to be too complex for the users. We believe that designing visual representations to answer each of these questions individually would be more effective and may reach better results.

Nevertheless, as discussed in the User Study section, participants' low performance not only resulted from the visual metaphor but also the expertise on RF models. Among the participants, few know the RF technique in detail, indicating that people with less expertise can use ExMatrix GE and LE/UR visualizations, but the LE/SC representation is more suitable for experts. In general, despite the complexity of the questions we ask participants to solve, they acknowledged the ExMatrix potential, expressing encouraging remarks, including ``\textit{... this solution ... allows a deeper understanding of how each particular rule or feature impacted on the final the decision/classification.}'' or ``\textit{I think the ExMatrix can be used in a variety of domains, from E-commerce to Healthcare...}''. 

%\subsection{Other Applications besides Random Forests}

Although we design ExMatrix with RF interpretability in mind, it can be readily applied to DT models, such as the ones used as surrogates for black-box models as Artificial Neural Networks and Support Vector Machines, or approaches based on logic rules such as Decision Tables since the core of our method is the visualization of rules. Another compelling scenario that can be explored is the engineering of models. In this case, through rule selection and filtering, simplified models could be derived where, for instance, only high coverage rules are employed or any other subset of interest. Also, model construction and improvement can be supported. The visual metaphors we propose can be easily applied to the analysis and comparison of RF models resulting from different parametrizations, such as different numbers of trees and their maximum depth. Therefore, allowing machine learning engineers to go beyond accuracy and error when building a model.

\section{Conclusions and Future Work}

In this paper, we present \textit{Explainable Matrix (ExMatrix)}, a novel method for Random Forest (RF) model interpretability. ExMatrix uses a matrix-like visual metaphor, where logic rules are rows, features are columns, and rules predicates are cells, allowing to obtain overviews of models (Global Explanations) and audit classification results (Local Explanations). Although simple, ExMatrix visual representations are powerful and support the execution of tasks that are challenging to perform without proper visualizations. To attest ExMatrix usefulness, we present one use-case and two hypothetical usage scenarios, showing that RF models can be interpreted beyond what is granted by usual metrics, like accuracy or error rate. Although our primary goal is to aid in RF models global and local interpretability, the ExMatrix method can also be applied for the analysis of Decision Trees, such as the ones used as surrogates models, or any other technique based on logic rules, opening up new possibilities for future development and use. We plan as future work to create new ordering and filtering criteria along with aggregation approaches to improve the current ExMatrix explanations and, more importantly, to conceive new ones. Another fascinating forthcoming work is creating optimized rule-based models from complex RF models, which we also intend to investigate.

%% if specified like this the section will be committed in review mode
\acknowledgments{The authors wish to thank the valuable comments and suggestions obtained from the reviewers, as well as the support received from the Qualification Program of the Federal Institute of S\~ao Paulo (IFSP). We acknowledge the support of the Natural Sciences and Engineering Research Council of Canada (NSERC).}

\bibliographystyle{abbrv}

\bibliography{ref}
\end{document}